\documentclass[sigconf, review]{acmart}
\renewcommand\footnotetextcopyrightpermission[1]{} 
\usepackage[utf8]{inputenc} 
\usepackage[T1]{fontenc}    
\usepackage{hyperref}       
\usepackage{url}            
\usepackage{booktabs}       
\usepackage{amsfonts}       
\usepackage{nicefrac}       
\usepackage{microtype}      

\setcopyright{none}

\usepackage[english]{babel}
\usepackage[utf8]{inputenc}
\usepackage{amsmath}
\usepackage{amssymb}
\usepackage{graphicx}
\usepackage[colorinlistoftodos]{todonotes}
\usepackage{eurosym}
\usepackage{hyperref}
\usepackage{subcaption}
\usepackage{natbib}
\usepackage{mathtools}
\usepackage{bbm}
\usepackage{wrapfig}
\DeclareMathOperator\arctanh{arctanh}
\usepackage{enumitem}
\setenumerate{leftmargin=4mm}
\setitemize{leftmargin=4mm}

\author{Benjamin Paul Chamberlain}
\orcid{}
\affiliation{
  \department{Department of Computing}
  \institution{Imperial College London}
}
\email{bpc14@imperial.ac.uk}

\author{James R Clough}
\orcid{}
\affiliation{
  \department{Centre for Complexity Science}
  \institution{Imperial College London}
}
\email{jrc309@imperial.ac.uk}

\author{Marc Peter Deisenroth}
\orcid{}
\affiliation{
  \department{Department of Computing}
  \institution{Imperial College London}
}
\email{m.deisenroth@imperial.ac.uk}

\date{\today}

\acmYear{2017}
\copyrightyear{2017}

\begin{document}

\title{Neural Embeddings of Graphs in Hyperbolic Space}

\begin{abstract}
Neural embeddings have been used with great success in Natural Language Processing (NLP). They provide compact representations that encapsulate word similarity and attain state-of-the-art performance in a range of linguistic tasks. The success of neural embeddings has prompted significant amounts of research into applications in domains other than language. One such domain is graph-structured data, where embeddings of vertices can be learned that encapsulate vertex similarity and improve performance on tasks including edge prediction and vertex labelling. For both NLP and graph based tasks, embeddings have been learned in high-dimensional Euclidean spaces.
However, recent work has shown that the appropriate isometric space for embedding complex networks is not the flat Euclidean space, but negatively curved, hyperbolic space. We present a new concept that exploits these recent insights and propose learning neural embeddings of graphs in hyperbolic space. We provide experimental evidence that embedding graphs in their \emph{natural} geometry significantly improves performance on downstream tasks for several real-world public datasets. 
\end{abstract}


\maketitle

\section{Introduction}
Embedding (or vector space) methods find a lower-dimensional continuous space in which to represent high-dimensional complex data \cite{Roweis2000, Belkin2001}.  The distance between objects in the lower-dimensional space gives a measure of their similarity. This is usually achieved by first postulating a low-dimensional vector space and then optimising an objective function of the vectors in that space. 
Vector space representations provide three principle benefits over sparse schemes: (1) They encapsulate similarity, (2) they are compact, (3) they perform better as inputs to machine learning models \cite{Salton1975}.
This is true of graph structured data where the native data format is the adjacency matrix, a typically large, sparse matrix of connection weights. 

Neural embedding models are a flavour of embedding scheme where the vector space corresponds to a subset of the network weights, which are learned through backpropagation.
Neural embedding models have been shown to improve performance in a large number of downstream tasks across multiple domains. These include word analogies~\cite{Mikolov2013, Mnih2013}, machine translation~\cite{Sutskever2014}, document comparison \cite{Kusner2015}, missing edge prediction~\cite{Grover}, vertex attribution~\cite{Perozzi2014}, product recommendations~\cite{Grbovic2015, Baeza-yates2015}, customer value prediction~\cite{Kooti2017, Chamberlain2017} and item categorisation~\cite{Barkan2016}. In all cases the embeddings are learned without labels (unsupervised) from a sequence of entities. 

To the best of our knowledge, all previous work on neural embedding models either explicitly or implicitly (by using the Euclidean dot product) assumes that the vector space is Euclidean. Recent work from the field of complex networks has found that many interesting networks, such as the Internet \cite{Boguna2010} or academic citations \cite{Clough2015a,Clough2016} can be well described by a framework with an underlying non-Euclidean hyperbolic geometry. Hyperbolic geometry provides a continuous analogue of tree-like graphs, and even infinite trees have nearly isometric embeddings in hyperbolic space \cite{Gromov}. Additionally, the defining features of complex networks, such as power-law degree distributions, strong clustering and hierarchical community structure, emerge naturally when random graphs are embedded in hyperbolic space~\cite{Krioukov}.

The starting point for our model is the celebrated word2vec Skipgram architecture, which is shown in Figure~\ref{fig:skipgram} \cite{Mikolov2013,Mikolov2013a}. Skipgram is a shallow neural network with three layers: (1) An input projection layer that maps from a one-hot-encoded to a distributed representation, (2) a hidden layer, and (3) an output softmax layer. The network is necessarily simple for tractability as there are a very large number of output states (every word in a language). 
Skipgram is trained on a sequence of words that is decomposed into (input word, context word)-pairs. The model employs two separate vector representations, one for the input words and another for the context words, with the input representation comprising the learned embedding. The word pairs are generated by taking a sequence of words and running a sliding window (the context) over them. As an example the word sequence ``chance favours the prepared mind'' with a context window of size three would generate the following training data: (chance, favours), (chance, the), (favours, chance), ... \}. Words are initially randomly allocated to vectors within the two vector spaces. Then, for each training pair, the vector representations of the observed input and context words are pushed towards each other and away from all other words (see Figure~\ref{fig:skipgram_updates}). 

The concept can be extended from words to network structured data using random walks to create sequences of vertices. The vertices are then treated exactly analogously to words in the NLP formulation. This was originally proposed as DeepWalk~\cite{Perozzi2014}. Extensions varying the nature of the random walks have been explored in LINE~\cite{Tang2015} and Node2vec~\cite{Grover}. 



\paragraph{Contribution}
In this paper, we introduce the new concept of neural embeddings in hyperbolic space. We formulate backpropagation in hyperbolic space and show that using the natural geometry of complex networks improves performance in vertex classification tasks across multiple networks.

\section{Hyperbolic Geometry}
Hyperbolic geometry emerges from relaxing Euclid's fifth postulate (the parallel postulate) of geometry. In hyperbolic space there is not just one, but an infinite number of parallel lines that pass through a single point. This is illustrated in Figure~\ref{fig:parallel} where every line is parallel to the bold, blue line and all pass through the same point. Hyperbolic space is one of only three types of isotropic spaces that can be defined entirely by their curvature. The most familiar is Euclidean, which is flat, having zero curvature. Space with uniform positive curvature has an elliptic geometry (e.g. the surface of a sphere), and space with uniform negative curvature is called hyperbolic, which is analogous to a saddle-like surface.
As, unlike Euclidean space, in hyperbolic space even infinite trees have nearly isometric embeddings, it has been successfully used to model complex networks with hierarchical structure, power-law degree distributions and high clustering \cite{Krioukov}.

One of the defining characteristics of hyperbolic space is that it is in some sense \textit{larger} than the more familiar Euclidean space; the area of a circle or volume of a sphere grows exponentially with its radius, rather than polynomially.
This suggests that low-dimensional hyperbolic spaces may provide effective representations of data in ways that low-dimensional Euclidean spaces cannot.
However this makes hyperbolic space hard to visualise as even the 2D hyperbolic plane can not be isometrically embedded into Euclidean space of any dimension,(unlike elliptic geometry where a 2-sphere can be embedded into 3D Euclidean space). 
For this reason there are many different ways of representing hyperbolic space, with each representation conserving some geometric properties, but distorting others. In the remainder of the paper we use the Poincar\'e disk model of hyperbolic space.

\subsection{Poincar\'e Disk Model}
 \begin{figure}[tb]
 \centering

 \begin{subfigure}[t]{0.4\hsize}
        \centering
        \includegraphics[width=\hsize]{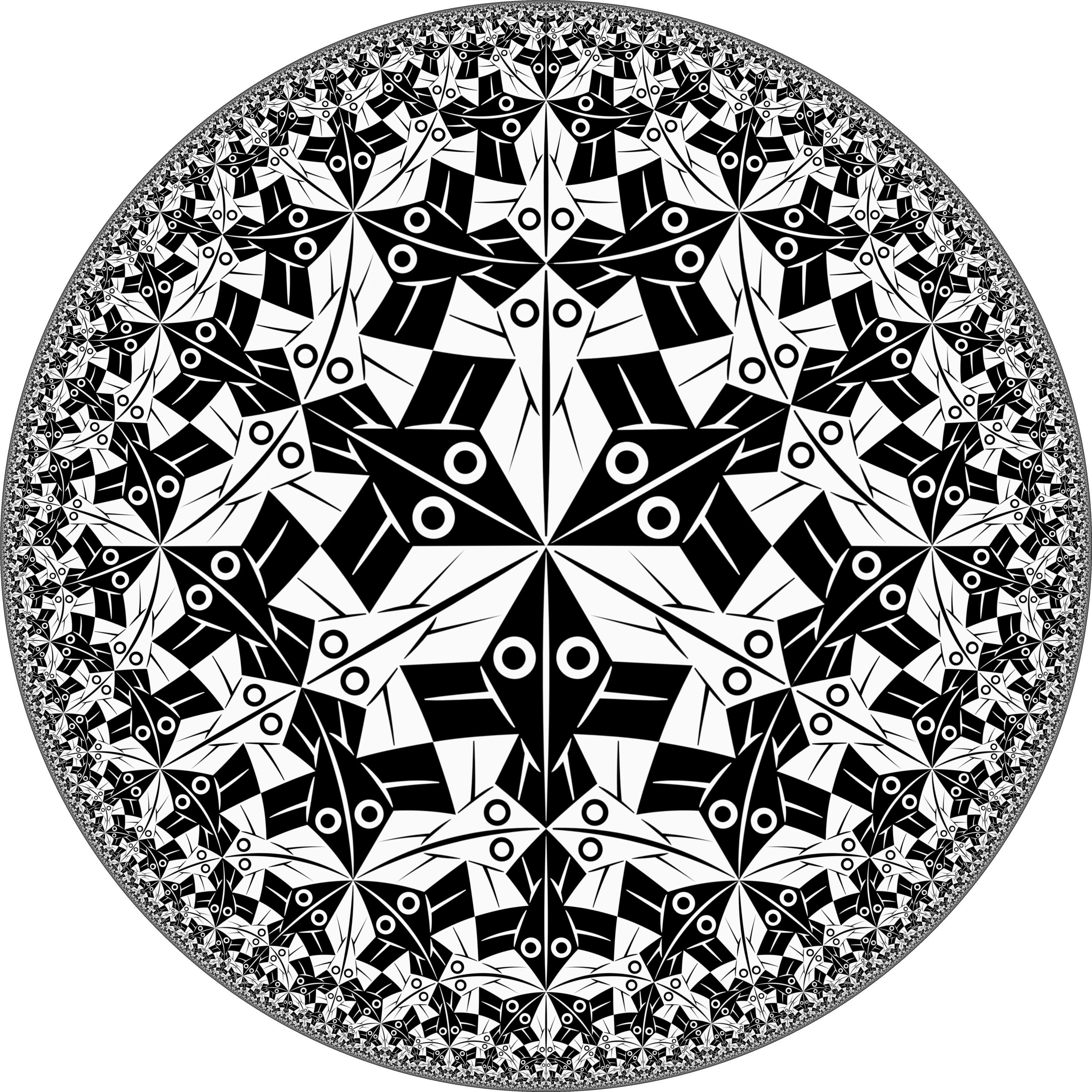}
        \caption{``Circle Limit 1'' by M.C. Escher illustrates the Poincar\'e disc model of hyperbolic space. Each tile is of constant area in hyperbolic space, but vanishes in Euclidean space at the boundary.}
        \label{fig:circle_limit1}
    \end{subfigure}%
   \hspace{10mm}
   \begin{subfigure}[t]{0.4\hsize}
        \centering
        \includegraphics[width=\hsize]{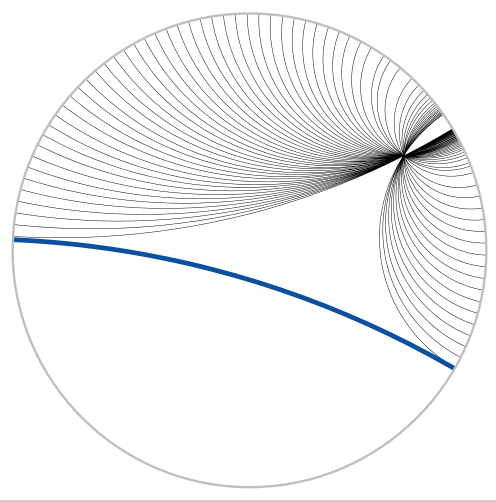}
        \caption{A set of straight lines in the Poincare disk that all pass through a given point and are all parallel to the blue (thicker) line.}
        \label{fig:parallel}
    \end{subfigure}%
    \caption{Illustrations of properties of hyperbolic space.  \subref{fig:circle_limit1} Tiles of constant area \subref{fig:parallel} Parallel lines.}
    \label{fig:hyperbolic illustration}
    \end{figure}

The Poincar\'e disk models two-dimensional hyperbolic space where the infinite plane is represented as a unit disk.
We work with the two-dimensional disk, but it is easily generalised to the $d$-dimensional Poincar\'e ball, where hyperbolic space is represented as a unit $d$-ball. 

In this model hyperbolic distances grow exponentially towards the edge of the disk. The circle's boundary represents infinitely distant points as the infinite hyperbolic plane is squashed inside the finite disk. This property is illustrated in Figure~\ref{fig:circle_limit1} where each tile is of constant area in hyperbolic space, but the tiles rapidly shrink to zero area in Euclidean space.
Although volumes and distances are warped, the Poincar\'e disk model is \textbf{conformal}, meaning that Euclidean and hyperbolic angles between lines are equal. 
Straight lines in hyperbolic space intersect the boundary of the disk orthogonally and appear either as diameters of the disk, or arcs of a circle. Figure~\ref{fig:parallel} shows a collection of straight hyperbolic lines in the Poincar\'e disk. Just as in spherical geometry, the shortest path from one place to another is a straight line, but appears as a curve on a flat map. Similarly, these straight lines show the shortest path (in terms of distance in the underlying hyperbolic space) from one point on the disk to another, but they appear curved. This is because it is quicker to move close to the centre of the disk, where distances are shorter, than nearer the edge.
In our proposed approach, we will exploit both the conformal property and the circular symmetry of the Poincar\'e disk.

Overall, the geometric intuition motivating our approach is that vertices embedded near the middle of the disk can have more close neighbours than they could in Euclidean space, whilst vertices nearer the edge of the disk can still be very far from each other.

\subsection{Inner Product, Angles, and Distances}
The mathematics is considerably simplified if we exploit the symmetries of the model and describe points in the Poincar\'e disk using polar coordinates, $x = (r_e,\theta)$, with $r_e \in [0, 1)$ and $\theta \in [0, 2\pi)$.
To define similarities and distances, we require an inner product. In the Poincar\'e disk, the \emph{inner product} of two vectors $x = (r_x, \theta_x)$ and $y=(r_y, \theta_y)$ is given by
\begin{align}
\langle x,y \rangle &= \Vert x \Vert \Vert y \Vert \cos (\theta_x - \theta_y) \\
&= 4 \arctanh r_x \arctanh r_y \cos(\theta_x - \theta_y) 
\label{eq:poincare_inner}
\end{align}
The \emph{distance} of $x = (r_e,\theta)$ from the origin of the hyperbolic co-ordinate system is given by $r_h = 2 \arctanh r_e$ and the circumference of a circle of hyperbolic radius R is $C = 2 \pi \sinh R$.

\section{Neural Embedding in Hyperbolic Space}

\begin{figure}[tb]
\centering
\includegraphics[width=0.48\textwidth]{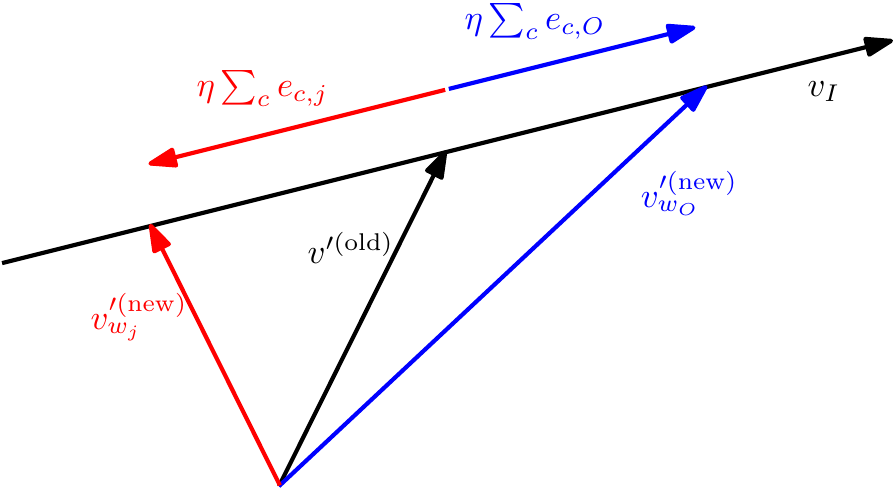}
\caption{Geometric interpretation of the update equations in the Skipgram model. The vector representation of the output vertex $v_{w_O}^{\prime(\mathrm{new})}$ is moved closer (blue) to the vector representation of the input vertex $v_I$, while all other vectors $v_{w_j}^{\prime(\mathrm{new})}$ move further away (red). The magnitude of the change is proportional to the prediction error.}
\label{fig:skipgram_updates}
\end{figure}

\begin{figure*}
 \begin{center}
    \includegraphics[width=\hsize]{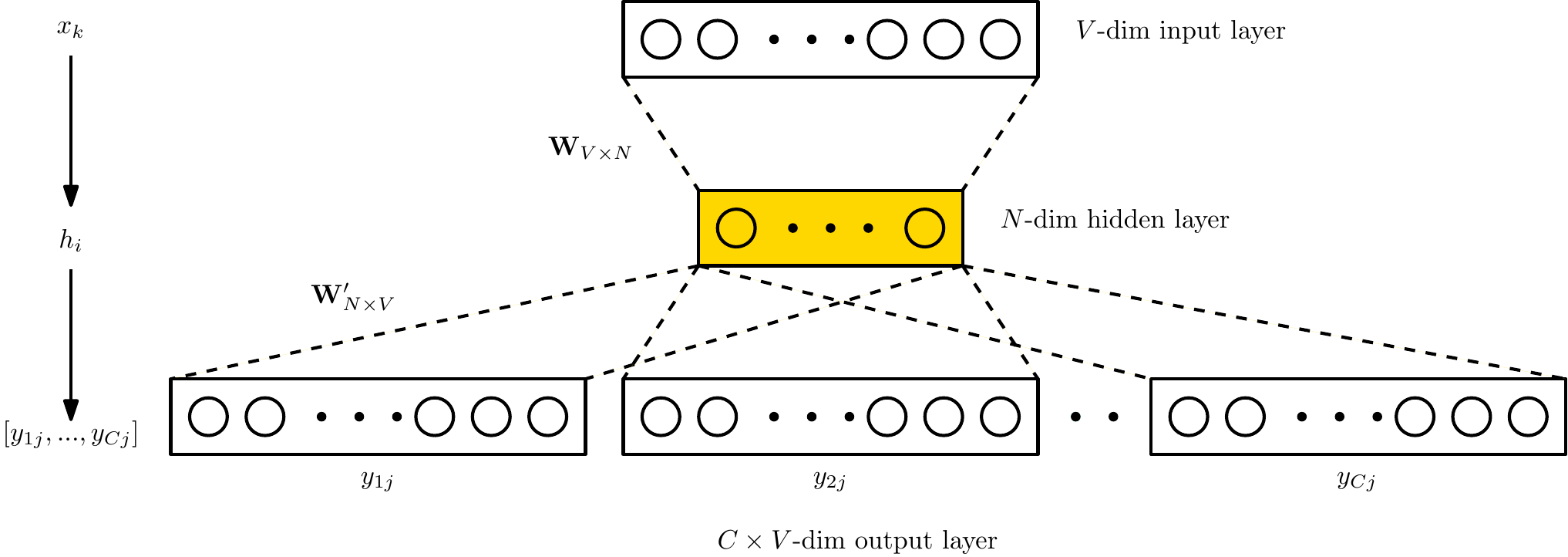}
  \end{center}
  \caption{The skipgram model predicts the context vertices from a single input vertex.}
  \label{fig:skipgram}
\end{figure*}

We adopt the original notation of \cite{Mikolov2013} whereby the input vertex is  $w_I$ and the output is $w_O$. Their corresponding vector representations are $v_{w_I}$ and $v'_{w_O}$, which are elements of the two vector spaces shown in Figure~\ref{fig:skipgram}, $\mathbf{W}$ and $\mathbf{W'}$ respectively.
Skipgram has a geometric interpretation, which we visualise in Figure~\ref{fig:skipgram_updates} for vectors in $\mathbf{W}^\prime$. Updates to $v^\prime_{w_j}$ are performed by simply adding (if $w_j$ is the observed output vertex) or subtracting (otherwise) an error-weighted portion of the input vector. Similar, though slightly more complicated, update rules apply to the vectors in $\mathbf{W}$. Given this interpretation, it is natural to look for alternative geometries in which to perform these updates.

To embed a graph in hyperbolic space we replace Skipgram's two Euclidean vector spaces ($\mathbf{W}$ and $\mathbf{W'}$ in Figure~\ref{fig:skipgram}) with two Poincar\'e disks. We learn embeddings by optimising an objective function that predicts output/context vertices from an input vertex, but we replace the Euclidean dot products used in Skipgram with hyperbolic inner products. A softmax function is used for the conditional predictive distribution
\begin{align}
p(w_O|w_I) = \frac{\exp (\langle v'_{w_O}, v_{w_I} \rangle)}{\sum_{i=1}^V\exp (\langle v'_{w_i}, v_{w_I} \rangle )}\,,
\label{eq:cond_dist}
\end{align}

where $v_{w_i}$ is the vector representation of the $i^{th}$ vertex, primed indicates members of the output vector space (See Figure~\ref{fig:skipgram}) and $\langle\cdot,\cdot \rangle$ is the hyperbolic inner product.
Directly optimising~\eqref{eq:cond_dist} is computationally demanding as the sum in the denominator extends over every vertex in the graph. Two commonly used techniques to make word2vec more efficient are (a) replacing the softmax with a hierarchical softmax~\cite{Mnih2008, Mikolov2013} and (b) negative sampling~\cite{Mnih2012, Mnih2013}. We use negative sampling as it is faster. 

\subsection{Negative Sampling}

Negative sampling is a form of Noise Contrastive Estimation (NCE)~\cite{Gutmann2012}. NCE is an estimation technique that is based on the assumption that a good model should be able to separate signal from noise using only logistic regression. 

As we only care about generating good embeddings, the objective function does not need to produce a well-specified probability distribution. The negative log likelihood using negative sampling is 
\begin{align}
E &= -\log \sigma (\langle v_{w_O}', v_{w_I} \rangle) -\hspace{-4mm} \sum_{w_j \in W_{neg}}\hspace{-2mm} \log \sigma(- \langle v_{w_j}', v_{w_I} \rangle) \\
&= -\log \sigma (u_O) - \sum_{j=1}^K \mathbb{E}_{w_j \sim P_n} [\log \sigma(- u_j)] 
\label{eq:loss}
\end{align}
where $v_{w_I}$, $v_{w_O}^\prime$ are the vector representation of the input and output vertices, $u_j = \langle v_{w_j}^\prime, v_{w_I}\rangle$, $W_{\rm neg}$ is a set of samples drawn from the noise distribution, $K$ is the number of samples and $\sigma$ is the sigmoid function. The first term represents the observed data and the second term the negative samples. To draw $W_{\rm neg}$, we specify the noise distribution $P_n$ to be unigrams raised to $\frac{3}{4}$ as in \cite{Mikolov2013}.

\subsection{Model Learning}

We learn the model using backpropagation. To perform backpropagation it is easiest to work in natural hyperbolic co-ordinates on the disk and map back to Euclidean co-ordinates only at the end. In natural co-ordinates $r \in (0,\infty)$, $\theta \in (0,2\pi]$ and $u_j = r_j r_I \cos(\theta_I - \theta_j)$. The major drawback of this co-ordinate system is that it introduces a singularity at the origin. To address the complexities that result from radii that are less than or equal to zero, we initialise all vectors to be in a patch of space that is small relative to its distance from the origin.

 The gradient of the negative log-likelihood in~\eqref{eq:loss} w.r.t. $u_j$ is given by
\begin{align}
\frac{\partial E}{\partial u_j} &= 
\begin{cases}
\sigma(u_j) - 1, & \text{if}\ w_j = w_O\\ 
\sigma(u_j), & \text{if}\ w_j = W_{neg}\\ 
0, & \text{otherwise}
\end{cases} 
\label{eq:error}
\end{align}
Taking the derivatives w.r.t. the components of vectors in $\mathbf{W'}$ (in natural polar hyperbolic co-ordinates)  yields
\begin{align}
\frac{\partial E}{\partial (r'_j)_k} &= \frac{\partial E}{\partial u_j}\frac{\partial u_j}{\partial (r'_j)_k} = \frac{\partial E}{\partial u_j} r_I \cos(\theta_I - \theta'_j) \\
\frac{\partial E}{\partial (\theta'_j)_k} & = \frac{\partial E}{\partial u_j} r_j'r_I \sin(\theta_I - \theta_j') \,.
\end{align}

The Jacobian is then
\begin{align}
\nabla  _\mathbf{r} E = \frac{\partial E}{\partial r} \mathbf{\hat{r}} + \frac{1}{\sinh r}\frac{\partial E}{\partial \theta} \mathbf{\hat{\theta}}\,,
\end{align}
which leads to
\begin{align}
r_j^{'new} &= 
\begin{cases} 
r_j^{'old} - \eta \epsilon_j r_I \cos(\theta_I - \theta'_j), & \text{if}\ w_j \in w_O\cup W_{neg}\\ 
r_j^{'old}, & \text{otherwise} 
\end{cases} \\
\theta_j^{'new} &= 
\begin{cases} 
\theta_j^{'old} - \eta \epsilon_j \frac{r_Ir_j}{\sinh{r_j}}\sin(\theta_I - \theta_j') , & \text{if}\ w_j \in w_O\cup W_{neg}\\ 
\theta_j^{'old}, & \text{otherwise} 
\end{cases} 
\end{align}
where $\eta$ is the learning rate and $\epsilon_j$ is the prediction error defined in Equation~\eqref{eq:error}. Calculating the derivatives w.r.t. the input embedding follows the same pattern, and we obtain
\begin{align}
\frac{\partial E}{\partial r_I} &= \sum_{j : w_j \in w_O \cup W_{neg}} \frac{\partial E}{\partial u_j}\frac{\partial u_j}{\partial r_I} \\
&= \sum_{j : w_j \in w_O \cup W_{neg}} \frac{\partial E}{\partial u_j} r_j' \cos(\theta_I - \theta'_j) \,,\\
\frac{\partial E}{\partial \theta_I} &= \sum_{j : w_j \in w_O \cup W_{neg}} \frac{\partial E}{\partial u_j}\frac{\partial u_j}{\partial \theta_I} \\
&= \sum_{j : w_j \in w_O \cup W_{neg}} -\frac{\partial E}{\partial u_j} r_I r_j' \sin(\theta_I - \theta'_j)\,. 
\end{align}
The corresponding update equations are
\begin{align}
r_I^{new} &= 
r_I^{old} - \eta \sum_{j : w_j \in w_O \cup W_{neg}} \epsilon_j r_j' \cos(\theta_I - \theta'_j)\,,\\ 
\theta_I^{new} &= 
\theta_I^{old} - \eta \sum_{j : w_j \in w_O \cup W_{neg}} \epsilon_j \frac{r_I r_j'}{\sinh r_I} \sin(\theta_I - \theta'_j)\,,
\end{align}
where $t_j$ is an indicator variable s.t. $t_j=1$ if and only if $w_j=w_O$, and $t_j = 0$ otherwise. On completion of backpropagation, the vectors are mapped back to Euclidean co-ordinates on the Poincar\'e disk through $\theta_h \to \theta_e$ and $r_h \to \tanh \frac{r_h}{2}$.


\section{Experimental Evaluation}
\begin{figure*}[t!]
    \centering
    \begin{subfigure}[t]{0.4\textwidth}
        \centering
        \includegraphics[width = \hsize]{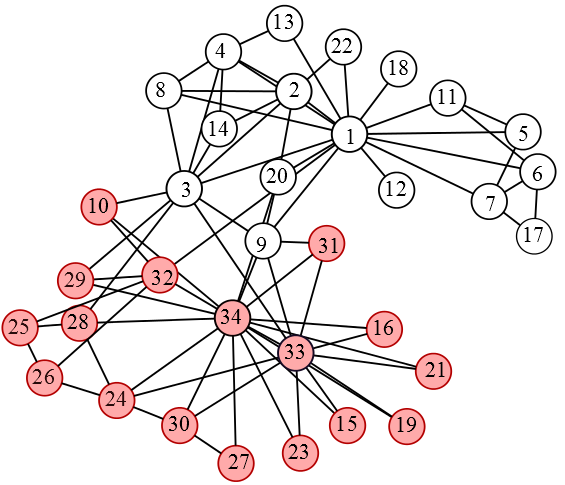}
        \caption{Zachary's karate network. The network is split into two factions.}
    \end{subfigure}%
    \\ 
    \begin{subfigure}[t]{0.48\textwidth}
        \centering
        \includegraphics[width = \hsize]{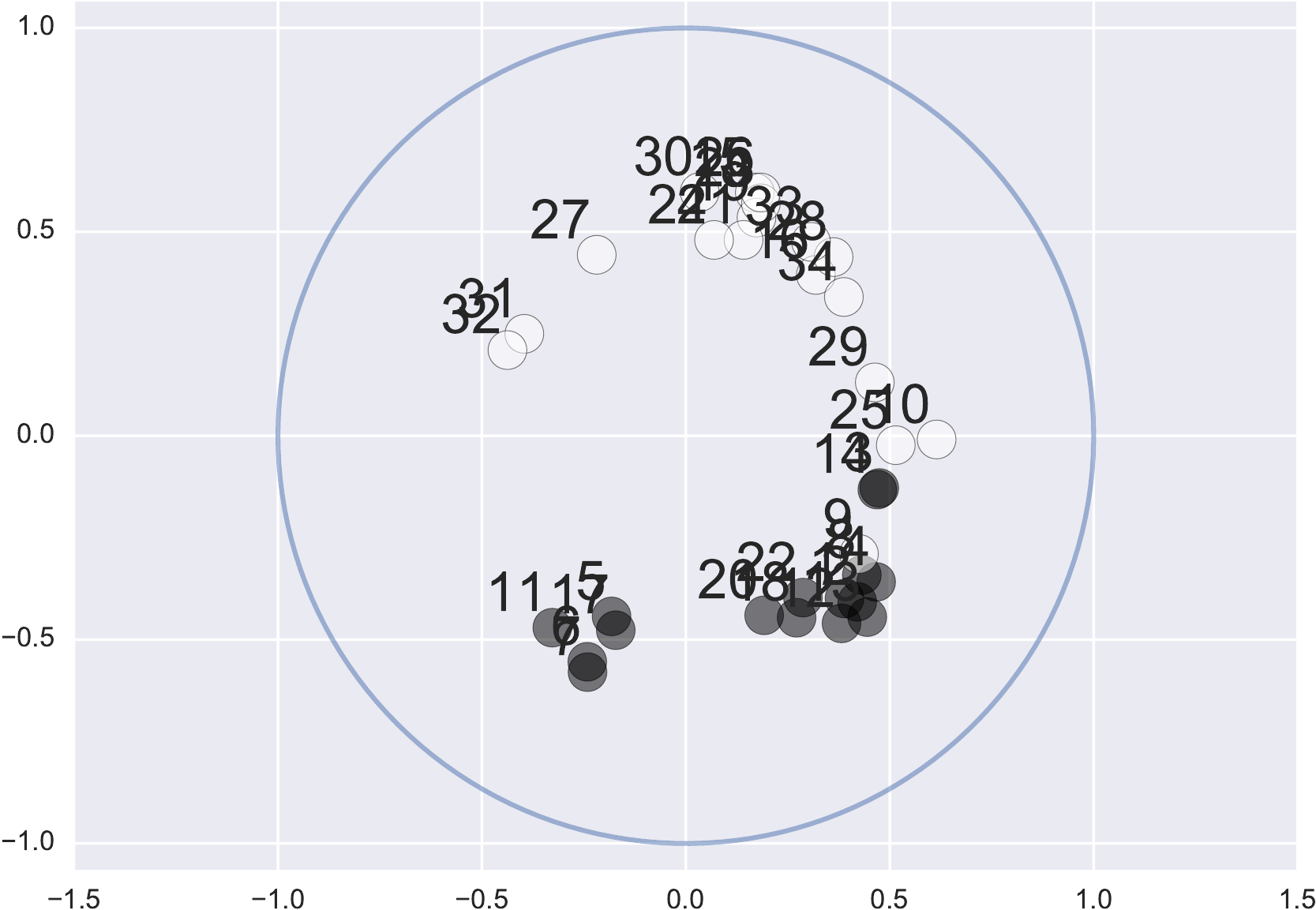}
        \caption{Two-dimensional hyperbolic embedding of the karate network in the Poincar\'e disk.}
    \end{subfigure}
        \hfill 
    \begin{subfigure}[t]{0.48\textwidth}
        \centering
        \includegraphics[width = \hsize]{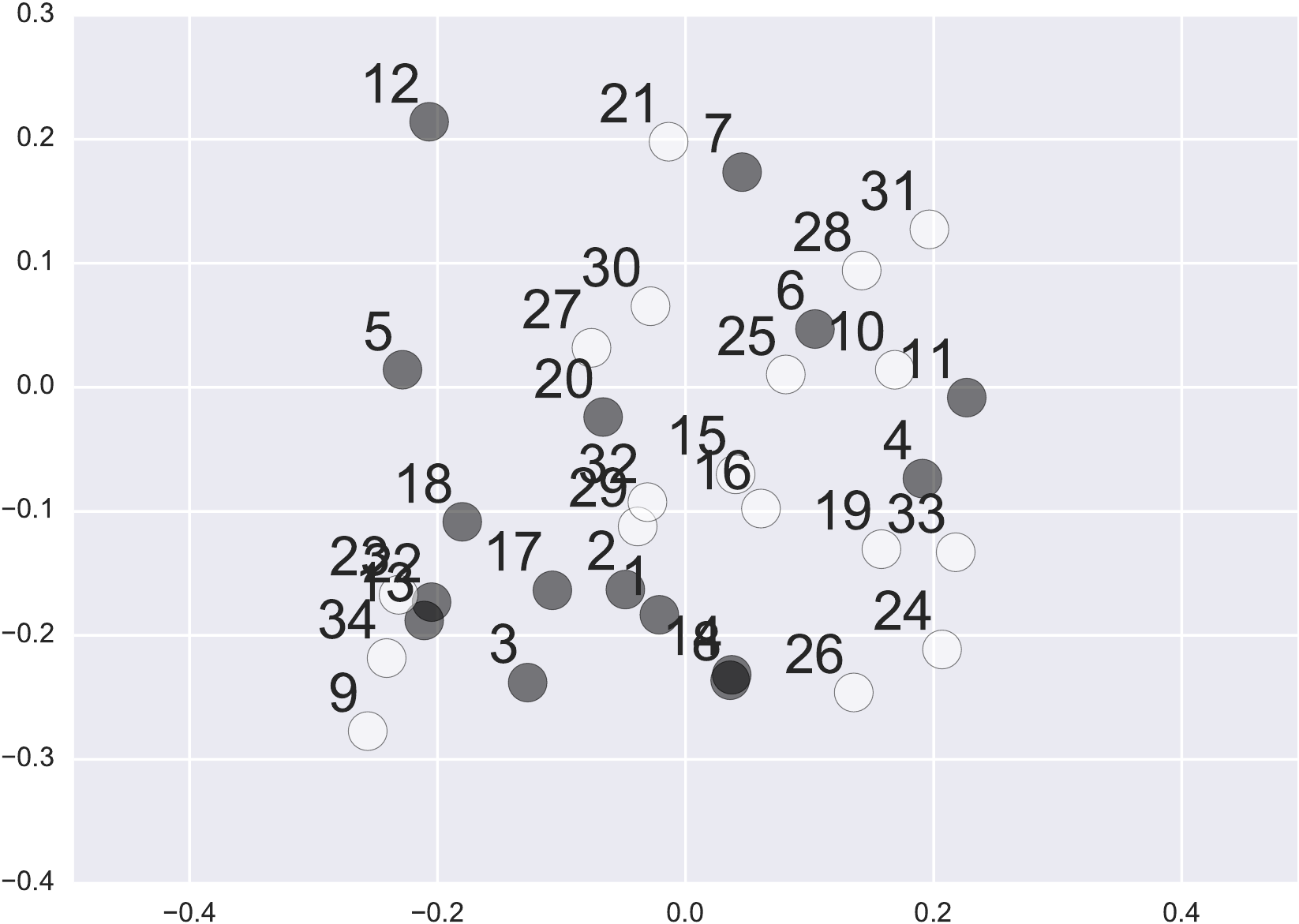}
        \caption{Two dimensional Deepwalk embedding of the karate network.}
    \end{subfigure}
    \caption{The factions of the  Zachary karate network are easily linearly separable when embedded in 2D hyperbolic space. This is not true when embedding in Euclidean space. Both embeddings were run for 5 epochs on the same random walks}
    \label{fig:embeddings}
\end{figure*}

In this section, we assess the quality of hyperbolic embeddings and compare them to embeddings in Euclidean spaces on a number of public benchmark networks.


\subsection{Datasets}
\begin{table}[tb]
  \centering
  \caption{Description of experimental datasets. `Largest class' gives the fraction of the dataset composed by the largest class and thereby provides the benchmark for random prediction accuracy.}
    \begin{tabular}{cllllc}
    \toprule
    name & |V| & |E| & |y| & largest class & Labels  \\
    \midrule
    karate     & 34 & 77 & 2 & 0.53 & Factions \\
    polbooks & 105 & 441 & 3 & 0.46 & Affiliation \\
    football & 115 & 613 & 12 & 0.11 & League \\
    adjnoun & 112 & 425 & 2 & 0.52 & Part of Speech \\
    polblogs & 1,224 & 16,781 & 2 & 0.52 & Affiliation \\
    \bottomrule
    \end{tabular}%
  \label{tab:datasets}%
\end{table}%
We report results on five publicly available network datasets for the problem of vertex attribution.
\begin{enumerate}
\item Karate: Zachary's karate club contains 34 vertices divided into two factions \cite{Zachary1977}.
\item Polbooks: A network of books about US politics published around the time of the 2004 presidential election and sold by the online bookseller Amazon.com. Edges between books represent frequent co-purchasing of books by the same buyers. 
\item Football: A network of American football games between Division IA colleges during regular season Fall 2000 \cite{Girvan2002}.
\item Adjnoun: Adjacency network of common adjectives and nouns in the novel David Copperfield by Charles Dickens \cite{Newman2006}.
\item Polblogs: A network of hyperlinks between weblogs on US politics, recorded in 2005 \cite{Adamic2005}.
\end{enumerate}
Statistics for these datasets are recorded in Table~\ref{tab:datasets}.

\begin{figure*}[h!]
    \centering
    \begin{subfigure}[t]{0.7\textwidth}
    \centering
    \includegraphics[width=\hsize]{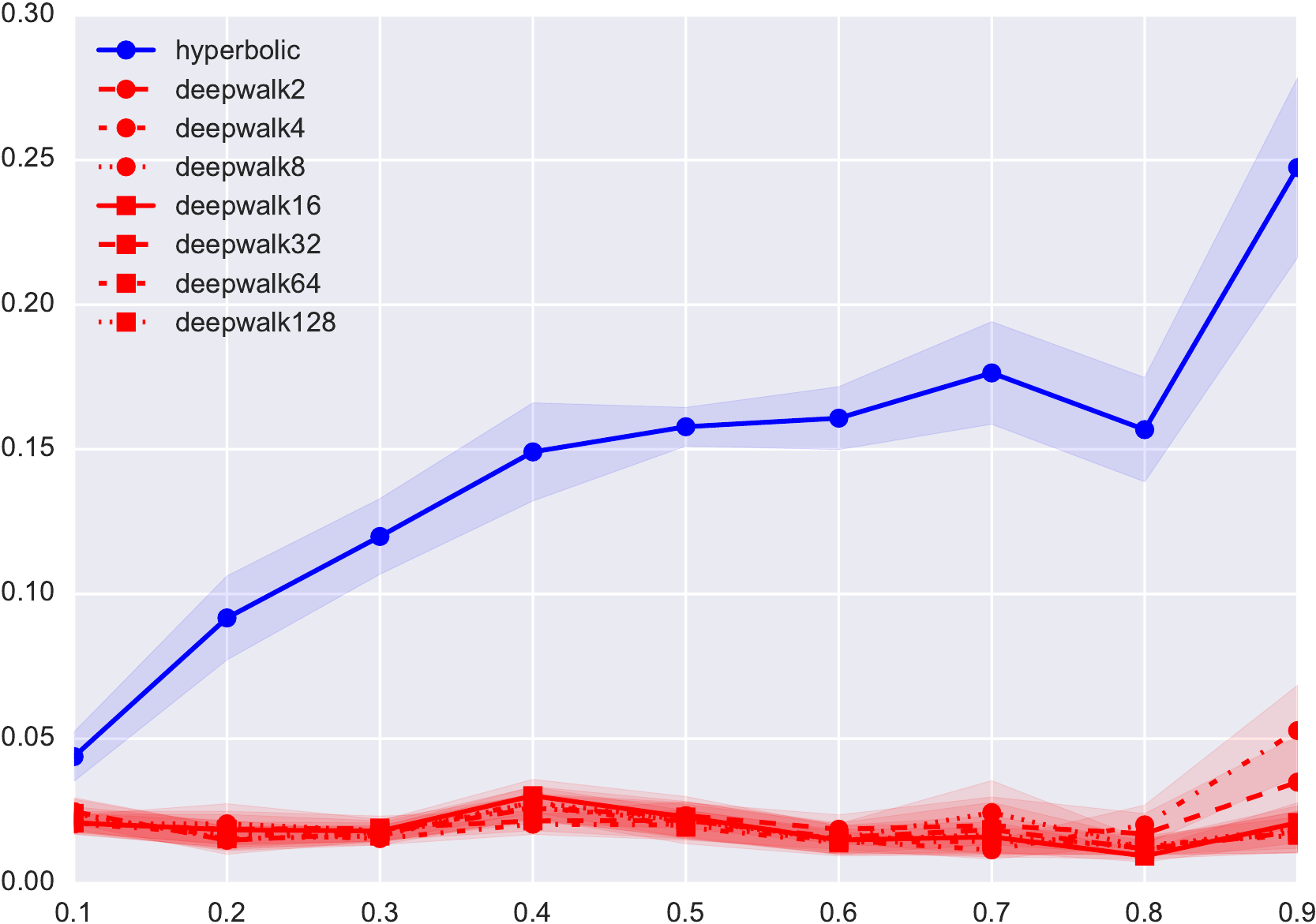}
    \caption{College football}
    \label{fig:football}
    \end{subfigure}
    \\ 
    \begin{subfigure}[t]{0.4\textwidth}
        \centering
        \includegraphics[width=\hsize]{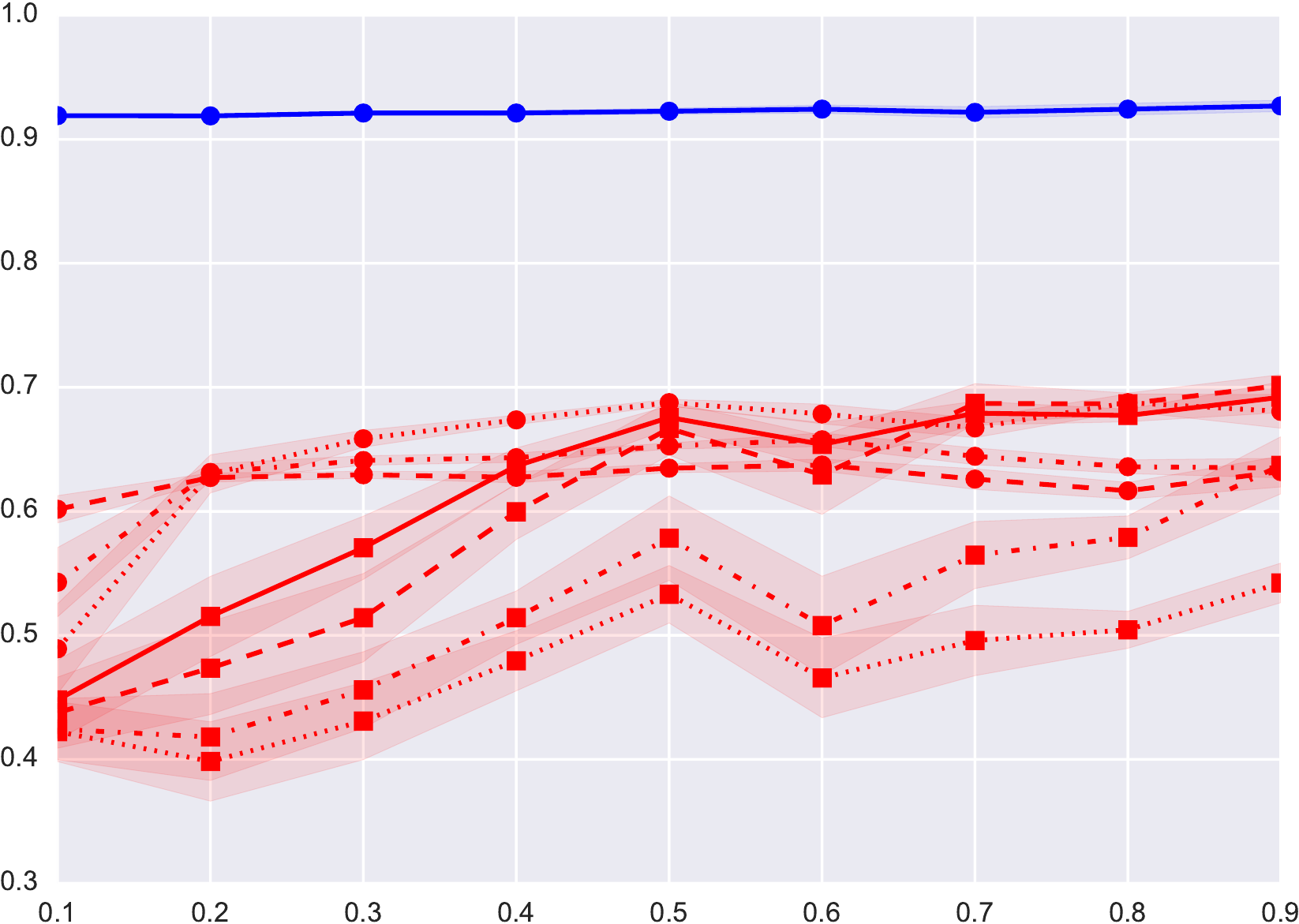}
        \caption{Political blogs.}
    \end{subfigure}%
   \hfill
    \begin{subfigure}[t]{0.4\textwidth}
        \centering
        \includegraphics[width=\hsize]{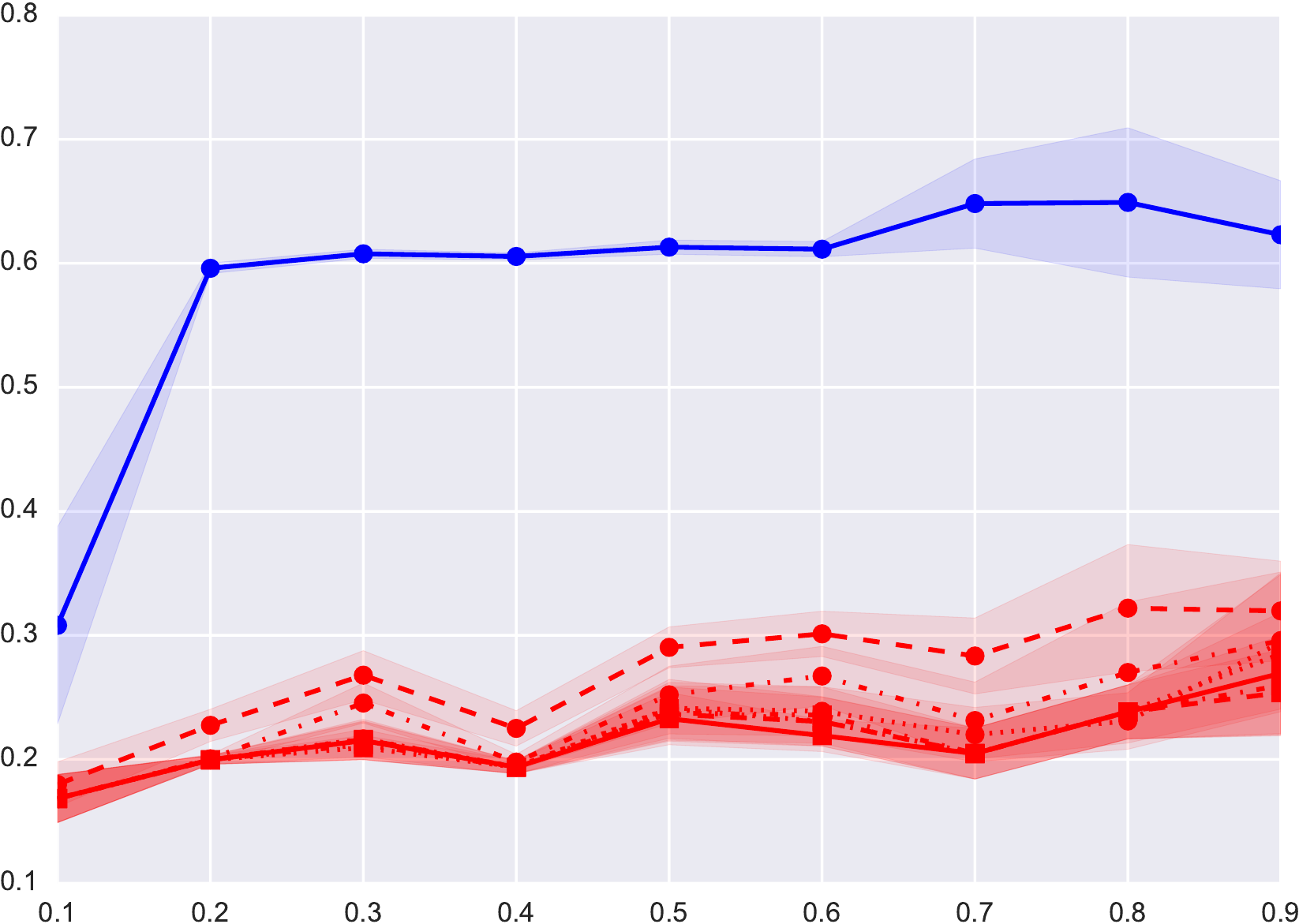}
        \caption{Political books.}
    \end{subfigure}
    \\
    \begin{subfigure}[t]{0.4\textwidth}
        \centering
        \includegraphics[width=\hsize]{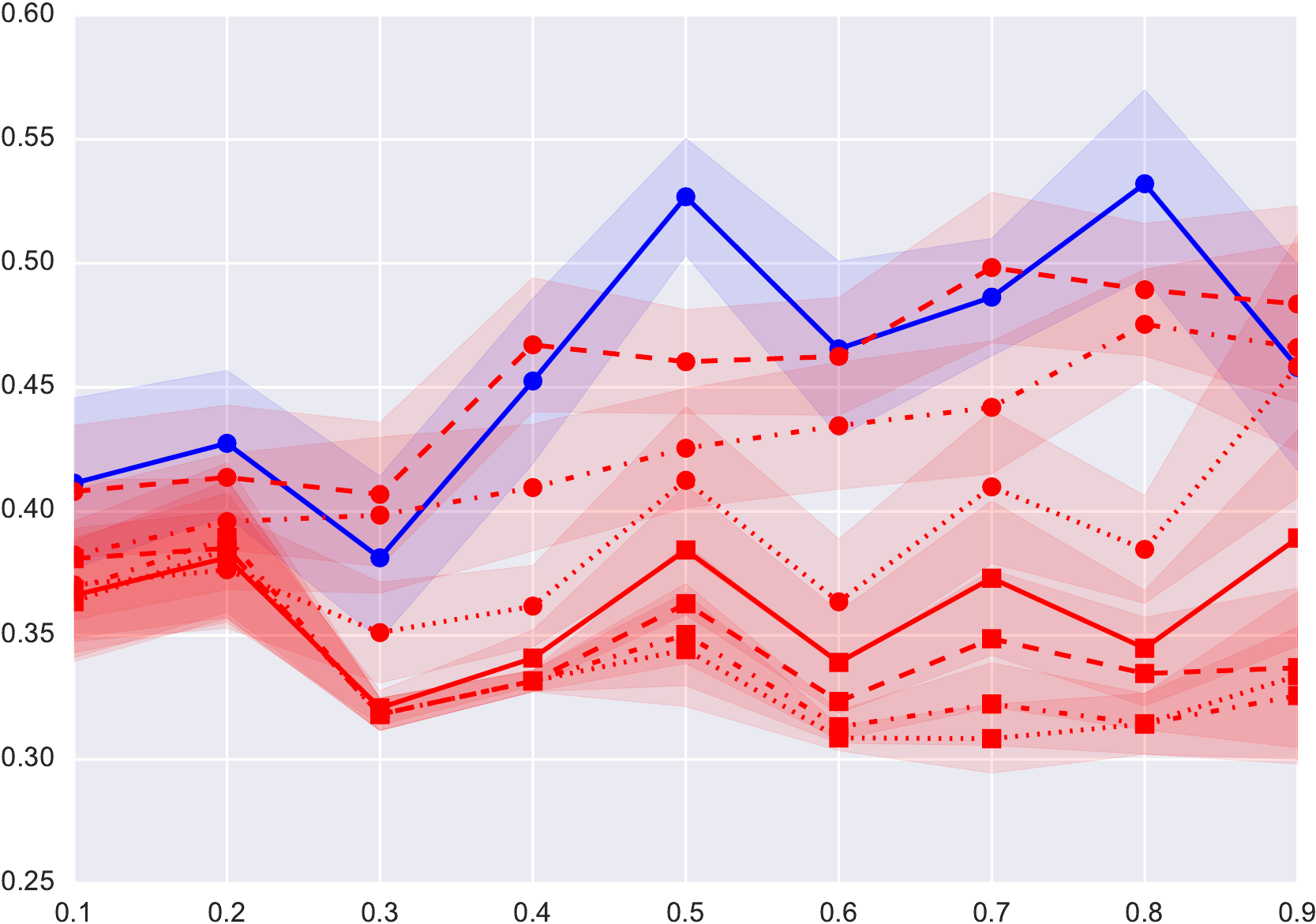}
        \caption{Word adjacencies.}
    \end{subfigure}
    \hfill
    \begin{subfigure}[t]{0.4\textwidth}
        \centering
        \includegraphics[width=\hsize]{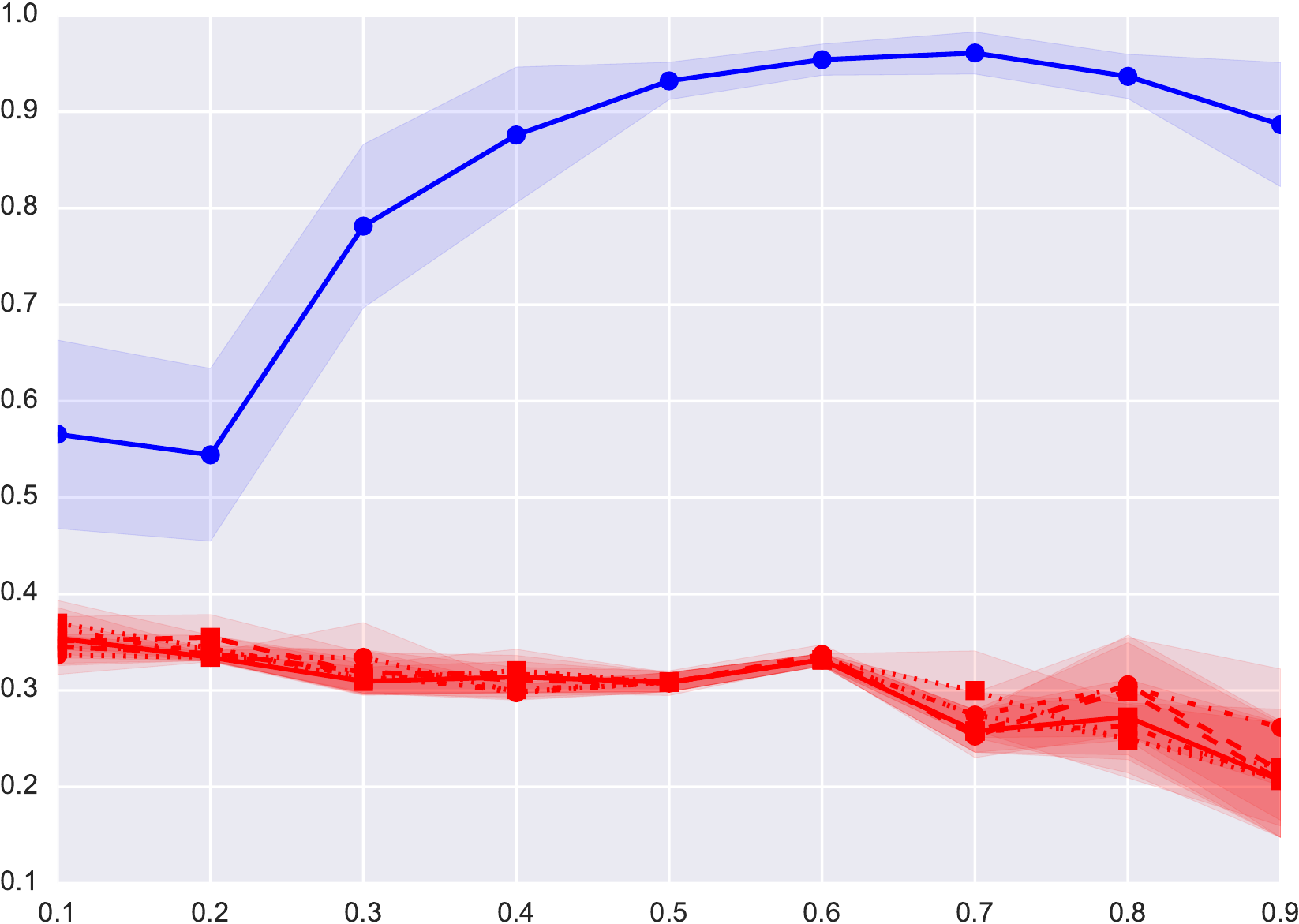}
        \caption{Karate network.}
    \end{subfigure}
    \caption{Macro F1 score ($y$-axis) against percentage of labelled vertices used for training ($x$-axis). In all cases hyperbolic embeddings (blue) significantly outperform Euclidean deepwalk embeddings (red). Error bars show standard error from the mean over ten repetitions. The legend used in subfigure (a) applies to all subfigures. A consistent trend across the datasets is that an embedding into a 2D hyperbolic space outperforms deepwalk architectures with embeddings ranging from 2D to 128D.}
    \label{fig:results}
\end{figure*}
\subsection{Visualising Embeddings}
To illustrate the utility of hyperbolic embeddings we compare embeddings in the Poincar\'e disk to the two-dimensional deepwalk embeddings for the 34-vertex karate network with two factions. The results are shown in Figure~\ref{fig:embeddings}. Both embeddings were generated by running for five epochs on an intermediate dataset of 34, ten step random walks, one originating at each vertex.

The figure clearly shows that the hyperbolic embedding is able to capture the community structure of the underlying network. When embedded in hyperbolic space, the two factions (black and white discs) of the underlying graph are linearly separable, while the Deepwalk embedding does not exhibit such an obvious structure.

\subsection{Vertex Attribute Prediction}

We evaluate the success of neural embeddings in hyperbolic space by using the learned embeddings to predict held-out labels of vertices in networks. In our experiments, we compare our embedding to deepwalk~\cite{Perozzi2014} embeddings of dimensions 2, 4, 8, 16, 32, 64 and 128. 
To generate embeddings we first create an intermediate dataset by taking a series of random walks over the networks. For each network we use a ten-step random walk originating at each vertex. 

The embedding models are all trained using the same parameters and intermediate random walk dataset. For deepwalk, we use the gensim~\cite{Rehurek2010} python package, while our hyperbolic embeddings are written in custom TensorFlow. In both cases, we use five training epochs, a window size of five and do not prune any vertices.

The results of our experiments are shown in Figure~\ref{fig:results}. The graphs show macro F1 scores against the percentage of labelled data used to train a logistic regression classifier. Here we follow the method for generating F1 scores when each test case can have multiple labels that is described in~\cite{Liu2006}. The error bars show one standard error from the mean over ten repetitions. The blue lines show hyperbolic embeddings while the red lines depict deepwalk embeddings at various dimensions. It is apparent that in all datasets hyperbolic embeddings significantly outperform deepwalk.

\section{Conclusion}

We have introduced the concept of neural embeddings in hyperbolic space. To the best of our knowledge, all previous embeddings models have assumed a flat Euclidean geometry. However, a flat geometry is not the natural geometry of all data structures.
A hyperbolic space has the property that power-law degree distributions, strong clustering and hierarchical community structure  emerge naturally when random graphs are embedded in hyperbolic space. It is therefore logical to exploit the structure of the hyperbolic space for useful embeddings of complex networks.  We have demonstrated that when applied to the task of classifying vertices of complex networks, hyperbolic space embeddings significantly outperform embeddings in Euclidean space.

\bibliographystyle{plain}
\bibliography{main.bbl}

\end{document}